\documentclass[10pt,twocolumn,letterpaper]{article}

\usepackage{cvpr}              % To produce the CAMERA-READY version
\usepackage[dvipsnames]{xcolor}

\usepackage{algorithm}
\usepackage{algorithmic}
\usepackage[accsupp]{axessibility}
\usepackage{booktabs}
\usepackage{multicol}
\usepackage{colortbl}
\usepackage{multirow}
\usepackage{amsthm}
\newtheorem{theorem}{Theorem} 
\newtheorem{definition}{definition}

\newtheorem{remark}{Remark}
\newtheorem{assumption}{Assumption}

\usepackage{amssymb}
\usepackage{pifont}
\usepackage{comment}
\usepackage[normalem]{ulem}
\usepackage{xcolor}

\definecolor{cvprblue}{rgb}{0.21,0.49,0.74}
\usepackage[pagebackref,breaklinks,colorlinks,citecolor=cvprblue]{hyperref}

\def\paperID{12} % *** Enter the Paper ID here
\def\confName{CVPR PBDL}
\def\confYear{2024}

\title{Imaging Signal Recovery Using Neural Network Priors \\ Under Uncertain Forward Model Parameters}

\author{ Xiwen Chen$^{1}$\footnotemark[1] \enspace  Wenhui Zhu$^{2}$\footnotemark[1]\enspace Peijie Qiu$^{3}$\footnotemark[1]\thanks{\textit{These authors contributed equally to this paper.}} 
\enspace Abolfazl Razi$^{1}$\thanks{\textit{Corresponding Author}}\\
$^{1}$ School of Computing, Clemson University\\ 
$^{2}$ School of Computing and Augmented Intelligence, Arizona State University\\ 
$^{3}$ McKeley School of Engineering, Washington University in St. Louis
}

\begin{document}
\maketitle

\begin{abstract}
% \vspace{-0.1cm}

% In this paper, we consider solving the inverse imaging problem (IIP) with uncertain forward model parameters, where an IIP is often ill-posed and relies on the prior knowledge. This is a common challenge in a variety of imaging tasks, where the forward model is parameterized (e.g. parameters of equipment), and using unmatched parameters of the forward model often fails to reconstruct the original signal. 

% \arr{This looks perfect!}

Inverse imaging problems (IIPs) arise in various applications, with the main objective of reconstructing an image from its compressed measurements. This problem is often ill-posed for being under-determined with multiple interchangeably consistent solutions. The best solution inherently depends on prior knowledge or assumptions, such as the sparsity of the image. Furthermore, the reconstruction process for most IIPs relies significantly on the imaging (i.e. forward model) parameters, which might not be fully known, or the measurement device may undergo calibration drifts. These uncertainties in the forward model create substantial challenges, where inaccurate reconstructions usually happen when the postulated parameters of the forward model do not fully match the actual ones. In this work, we devoted to tackling accurate reconstruction under the context of a set of possible forward model parameters that exist. Here, we propose a novel Moment-Aggregation (MA) framework that is compatible with the popular IIP solution by using a neural network prior. Specifically, our method can reconstruct the signal by considering all candidate parameters of the forward model simultaneously during the update of the neural network. We theoretically demonstrate the convergence of the MA framework, which has a similar complexity with reconstruction under the known forward model parameters. Proof-of-concept experiments demonstrate that the proposed MA achieves performance comparable to the forward model with the known precise parameter in reconstruction across both compressive sensing and phase retrieval applications, with a PSNR gap of 0.17 to 1.94 over various datasets, including MNIST, X-ray, Glas, and MoNuseg. This highlights our method's significant potential in reconstruction under an uncertain forward model.

%\arr{The above para is good but the flow is not easy to follow, you can start by saying "IIP arises in various fields where the goal is to reconstruct the image from its compressed measurements, and then mention it is all posed and hence relies on some prior knowledge or assumptions like the sparsity of image, and then mention that the reconstruction process in most (or a class of IIPs) depends on the imaging parameter which might not be fully known or the measurement device may go out of calibrations, etc. Then, This work is devoted to tackling this challenge ......". Also, explicitly mention that you are considering a case in which the parameters are within some range, or if you have a set of candidates for it, etc. }
\end{abstract}

% \vspace{-0.1cm}
\section{Introduction}
%\vspace{-0.1cm}
Inverse imaging problems (IIPs) aim to reconstruct a sought-after image $\boldsymbol{x}_0 \in \mathbb{R}^n$ from its measurements $\boldsymbol{y} \in \mathbb{R}^m$, where $m$ is often much smaller than $n$ and the observation is typically contaminated by some sort of observation noise $\boldsymbol{\eta}$. We have
\begin{equation}
\boldsymbol{y}=\mathcal{A}\left(\boldsymbol{x}_0;\theta^*\right)+\boldsymbol{\eta},
\end{equation}
where $\mathcal{A}(\cdot)$ denotes the forward imaging model, which is typically governed by different mathematical and physical principles and often parameterized by $\theta^*$. 
%Representative forward models are 
Some real-world examples of this paradigm include 
magnetic resonance imaging (MRI) \cite{lustig2008compressed,feng2016xd,leynes2024scan}, tomographic imaging \cite{yu2009compressed,lan2021compressed}, lensless photography \cite{monakhova2021untrained}, microscopic imaging \cite{zhang2018twin,li2020deep,chen2023dh}, and even image processing \cite{wang2014nonlinear,oliveri2017compressive,li2023nonlocal}, each of which with its own forward modeling stemmed from the underlying physics and utilized technology. 

% These observations  $\boldsymbol{y} \in \mathbb{R}^m$ are obtained from the unknown real data $\boldsymbol{x}_0\in \mathbb{R}^n$ by a forward process $\mathcal{A}(\cdot)$ that presented as
% % \begin{align}
% $\boldsymbol{y}=\mathcal{A}\left(\boldsymbol{x}_0\right)+\boldsymbol{\eta}$,
% % \end{align}
% where $\boldsymbol{\eta}$ is a noise term. 

\begin{figure}[!t]
% \vspace{-0.4cm}
\centering
\includegraphics[width=1\columnwidth]{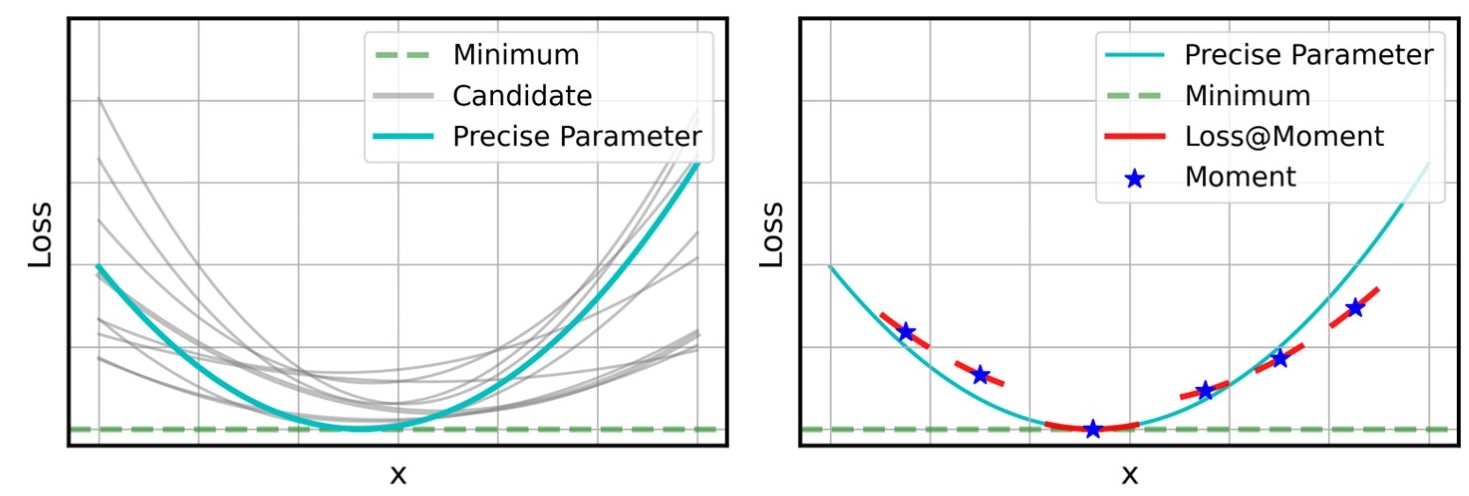} % Reduce the figure size so it is slightly narrower than the column. Don't use precise values for figure width. This setup will avoid overfull boxes.
%\vspace{-0.8cm}
\caption{The illustration of the Moment-Aggregation (MA) framework for IIPs with a neural network that considers the effect of all possible candidate parameters of the forward model simultaneously. MA loss is constructed after every forward propagation (we call this time point a "moment") and then is used to update parameters in backward propagation. \textbf{Left:} The losses by candidate forward model parameters, and one of them is the precise parameter. Their labels are unknown (i.e. the precise or not precise) during training. \textbf{Right:} The loss at different moments by MA. The loss is moment-wise convex$/$smooth, and the overall training can achieve the global minima as reconstruction using the precise parameter.  }
\label{fig:fig1}
% \vspace{-1cm}
\end{figure}

\begin{figure}[!t]
% \vspace{-0.4cm}
\centering
\includegraphics[width=0.9\columnwidth]{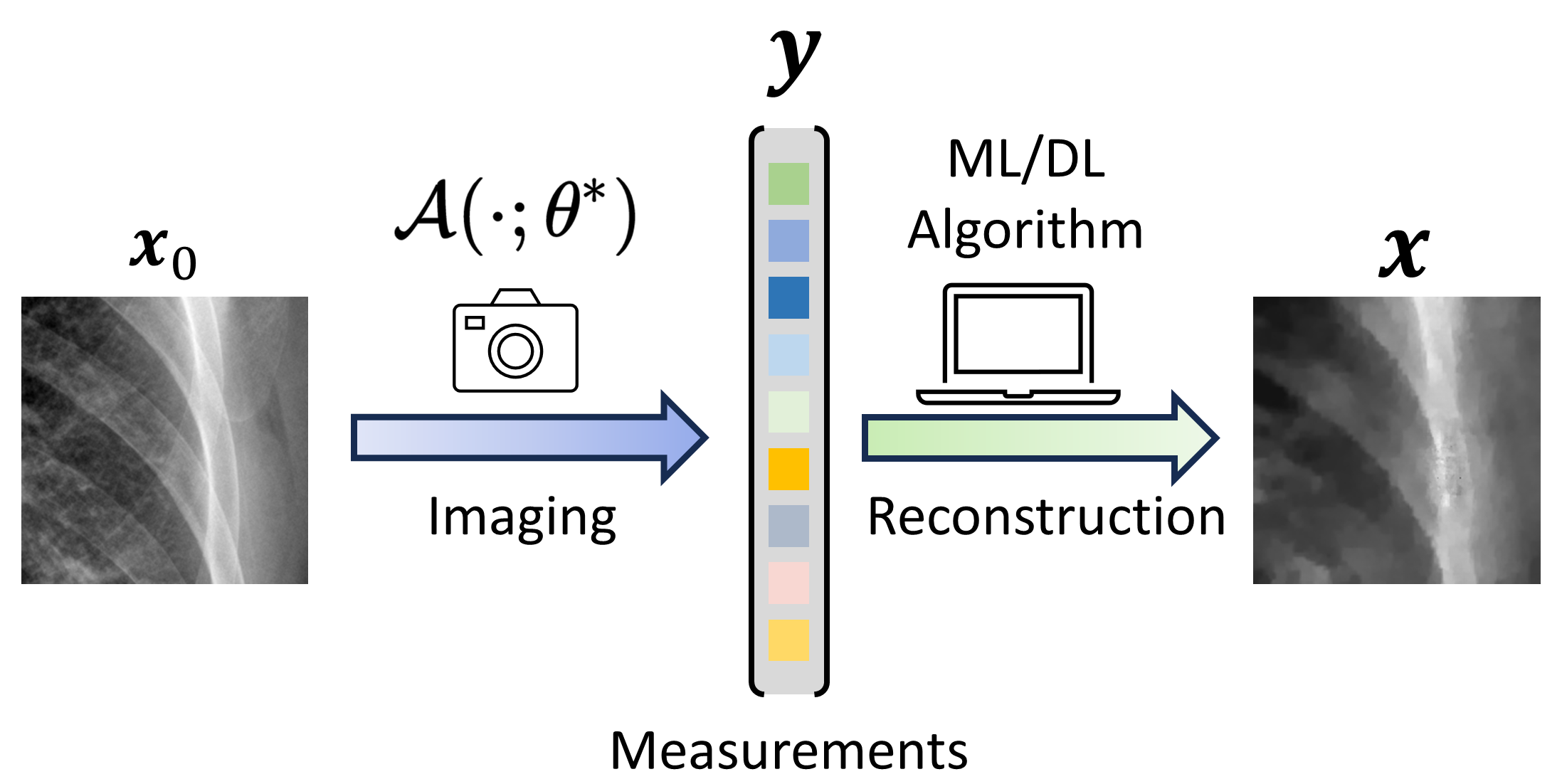} % Reduce the figure size so it is slightly narrower than the column. Don't use precise values for figure width. This setup will avoid overfull boxes.
%\vspace{-0.8cm}
\caption{The typical workflow of IIPs. First, a forward model is applied to a signal to obtain the measurement. Then the measurement is used to reconstruct the original signal via machine learning (ML) or deep learning (DL) algorithms.    }
\label{fig:quesiton}
% \vspace{-1cm}
\end{figure}

IIPs are typically ill-posed (underdetermined for $m<n$), which means they have multiple interchangeably consistent solutions. The core idea to solve these problems is incorporating prior information about the original signal (e.g., prior distribution, smoothness, sparsity, etc.) into the reconstruction algorithm. 
%This helps to reduce the search space and enhance the yielded reconstruction 
This enhances the reconstruction quality by reducing the search space and steering the algorithm toward the most probable and reality-compliant solution \cite{qayyum2022untrained}. Mathematically, an IIP is typically given in a variational formulation:
\begin{align}\label{eq:problem}
    \underset{\boldsymbol{x}}{\arg \min } \frac{1}{2}\|\boldsymbol{y}-\mathcal{A}(\boldsymbol{x};\theta^*)\|_2^2+\lambda_0 R(\boldsymbol{x}),
\end{align}
where $\boldsymbol{x}$ denotes the reconstructed image, $ R(\boldsymbol{x})$ denotes the regularization term governed by prior knowledge, and $\lambda_0$ controls the regularization strength. The typical workflow is shown in Fig. \ref{fig:quesiton}.

% these forward models, also known as forward processes, are application-specific, and most of them depend on knowledge of mathematics and physics. For example, compressive sensing (CS) uses random Gaussian matrix \cite{candes2006robust} while optical imaging uses angular spectrum method \cite{zhang2018twin}. However,

\begin{figure*}[]
% \vspace{-0.4cm}
\centering
\includegraphics[width=1\textwidth]{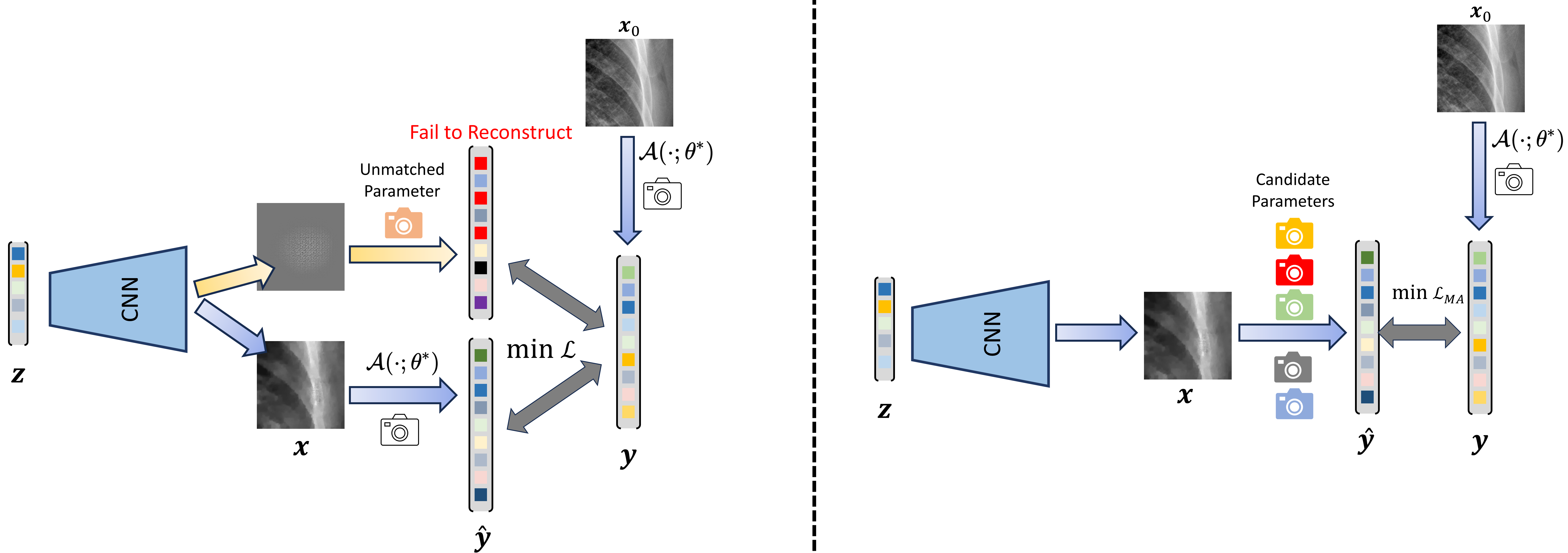} % Reduce the figure size so it is slightly narrower than the column. Don't use precise values for figure width. This setup will avoid overfull boxes.
% \vspace{-0.8cm}
\caption{Using CS-DIP to reconstruct $\boldsymbol{x}_0$ with measurement $\boldsymbol{y}$. \textbf{Left:} The signal is successfully reconstructed under the forward model with the precise parameter while failing under a wrong parameter. \textbf{Right:} Our method can successfully reconstruct $\boldsymbol{x}_0$ by optimizing under a set of candidate parameters (we assume one of them is close to the precise parameter).  }
\label{fig:fig2}
% \vspace{-1cm}
\end{figure*}

%It is noteworthy that one crucial issue for IIPs is that the effectiveness of signal reconstruction can be severely suppressed 
It is worth mentioning that one key issue of IIPs is that the quality of signal reconstruction can be severely declined
if the designed and implemented parameters of forward models do not match. Fig.\ref{fig:fig2}(\textbf{Left}) shows reconstruction using a forward model with the known precise parameter can successfully recover the signal while with a wrong parameter fails. 
This issue is general when recording microscopic images with low-cost equipment. The small scale and precision limitation of such equipment makes it challenging to accurately depict the forward model. Furthermore, another application scenario involves employing diverse setup parameters to capture various samples, wherein, due to an inadvertent mix-up or loss of the setup records, the forward model aligning with corresponding samples is necessary for accurate reconstruction. This process of rematching different setup configurations for distinct samples is recognized as a laborious and time-consuming endeavor, which is widely neglected by existing methods.
% due to constraints in computational resources and has not been addressed.

To address this issue, we consider several possible candidate parameters of forward models, and we formulate the recovery task under uncertain parameters of a forward model as a two-variable optimization problem. We propose a general optimization framework named Moment-Aggregation (MA) that is compatible with the state-of-the-art method for IIPs based on untrained neural network priors. Here, the moment is defined as the time point after forward propagation and before backward propagation. Aggregation means considering the effects of all possible candidates simultaneously (shown in Fig.\ref{fig:fig2}(\textbf{Right})). By using the gradient-stopping trick, we construct aggregation functions that are able to adjust according to the training process automatically. Subsequently, leveraging the advantages of neural network-based back-propagation for optimization, our framework can achieve a recovery accuracy comparable to the signal recovered by using the known precise parameter. An exemplary loss surface is shown in Fig. \ref{fig:fig1}.

In summary, our contribution is two-fold: i) We propose Moment-Aggregation, a general framework to solve IIPs under uncertainty parameters of the forward model. ii) We provide a theoretical analysis of MA. The experiments conducted on two applications, including compressive sensing and phase retrieval, confirm the feasibility of our method.

% In this paper, we study the most well-known under-determined linear measurement systems, termed compressive sensing (CS) \cite{}, which has been used in several imaging applications, such as magnetic resonance imaging (MRI) \cite{lustig2008compressed,feng2016xd}, tomography \cite{yu2009compressed}, lensless photography \cite{monakhova2021untrained}, and image processing \cite{wang2014nonlinear,oliveri2017compressive}. 

\section{Related Work}
\noindent\textbf{IIPs by Neural Network Priors.} The conventional methods to solve IIPs rely on handcrafted prior domain knowledge; however, these methods are often sensitive to the hyperparameters (e.g., $\lambda_0$ in Eq. \ref{eq:problem}. Note this is different from the parameter of the forward model) and often yield a poor recovery performance \cite{qayyum2022untrained}. Recent deep-learning methods, such as supervised learning \cite{he2016deep} and unsupervised learning \cite{zhu2017unpaired,chen2020simple}, demonstrate an outstanding ability to solve several image tasks. Due to this powerful tool, authors in \cite{shah2018solving,ulyanov2018deep,jalal2021robust,nguyen2022provable} show that inverse problems can be solved by using the prior from pre-trained generative models, which is known as learned network prior.  Along with the prior that is learned by massive training data, recently, the community \cite{ulyanov2018deep,gandelsman2019double} has observed that even without training on any dataset, the randomly initialized convolutional neural networks (CNNs) already hold the prior for image signals. This prior, often known as deep image prior (DIP), states that CNNs are able to capture a significant amount of low-level image statistics before any training on a specific image dataset. Hence, DIP becomes the natural choice to serve as the prior in IIPs (i.e. $R(\boldsymbol{x})$ in Eq. \ref{eq:problem}) and is employed by numerous works \cite{van2018compressed,li2020deep,kafle2021one,lan2021compressed,chen2022unsupervised,li2020deep,bai2021dual,chen2023dh,leynes2024scan,li2023nonlocal}. These works often involve a randomly initialized CNN-based generative model and solve the inverse
problem via training the network parameters. As these works often assume knowing the precise or near-precise parameter of the forward model, our work is orthogonal but complementary to them and aims to recover signals under a set of candidate parameters.    

% However, collecting application-specific datasets is often time-cost and may not be feasible for some applications due to privacy issues. For example,  

% \cite{hegde2018algorithmic,jagatap2019algorithmic,heckel2020compressive,scarlett2022theoretical,nguyen2022provable}

\noindent\textbf{Convergence Guarantee.} There are numerous works \cite{scarlett2022theoretical} provide the convergence and error guarantee for IIPs with neural network prior. For example, authors in \cite{bora2017compressed} prove a near-linear convergence rate for a $L-$Lipschitz continues generative network. Afterward, authors in \cite{hegde2018algorithmic} investigate the convergence rate by projected gradient descent with generative network prior, while authors in \cite{nguyen2022provable} study an algorithm based on Langevin dynamics with learned network prior. Likewise, with untrained network prior, authors in \cite{jagatap2019algorithmic} prove the convergence rate for under-parameterized networks and authors in \cite{heckel2020compressive} prove it for the over-parameterized networks.

It is observed that in order to ensure the derivation is tractable, these works often employ multiple assumptions, such as Lipschitz continues, the range of neural network, and the network only has linear layers and Relu activation functions \cite{scarlett2022theoretical}. We admit these assumptions simplify the real optimization process of the reconstruction, but their theoretical results offer enough insights for the community to develop further works. Again, their works often assume the forward model parameter is known, and in this work, we build theoretical analysis in our scenarios based on their conclusion.

% untrained neural network
% jagatap2019algorithmic ,heckel2020compressive,  

\noindent\textbf{Recovery with Uncertainty in CS.} There are several works \cite{rosenbaum2010sparse,rosenbaum2013improved,razi2019bayesian,le2023subgroup} studying the problem of mismatch measurement in CS. However, they often assume the error of the forward model is white additive noise and relatively small to the precise parameter. Besides, their theoretical guarantee is often designed for CS problems and relies on the Gaussianity assumption, which is difficult to generalize to broader scenarios. More importantly, authors in \cite{hegde2018algorithmic,jalal2021robust} show using neural network prior is relatively robust to such a noisy forward model. Contrastingly, we consider reconstruction with a discrete set of parameter candidates, and the distance among different measurements resulting from these forward models can be arbitrarily large.

% \begin{figure}[!t]
% % \vspace{-0.4cm}
% \centering
% \includegraphics[width=0.5\textwidth]{fig2.jpg} % Reduce the figure size so it is slightly narrower than the column. Don't use precise values for figure width. This setup will avoid overfull boxes.
% % \vspace{-0.8cm}
% \caption{ }
% \label{fig:fig2}
% % \vspace{-1cm}
% \end{figure}

\section{Problem Formulation}
Consider an observation/measurement $\boldsymbol{y}$ obtained by applying a forward model with a known parameter $\mathcal{A}(\cdot;\theta^*)$ to ground truth data $\boldsymbol{x}_0$, presented as,
\begin{align}\label{eq:problem_me}
\boldsymbol{y}=\mathcal{A}\left(\boldsymbol{x}_0;\theta^*\right)+\boldsymbol{\eta}.
\end{align}
Our problem now is to recover the signal from $\boldsymbol{y}$ and a set of candidate forward model parameters $ \Theta=\{ \theta_1,\cdots,\theta_{n_c} \}$, where $n_c$ denotes the total number of candidates. For simplicity, following \cite{nguyen2022provable}, we consider zero measurement noise, i.e. $\boldsymbol{\eta}=\boldsymbol{0}$. The objective function is now presented as,
\begin{align}\label{eq:objective_me}
    % \underset{\boldsymbol{x},i}{\arg \min }\quad F(\boldsymbol{x};\theta_i)= \frac{1}{2}\|\boldsymbol{y}-\mathcal{A}(\boldsymbol{x};\theta_i)\|_2^2,
    \quad F(\boldsymbol{x};\theta_i)= \frac{1}{2}\|\boldsymbol{y}-\mathcal{A}(\boldsymbol{x};\theta_i)\|_2^2,
\end{align}
where $\theta_i\in \Theta$. We omit the term $\lambda_0 R(\boldsymbol{x})$ as the prior is included in the neural network. 
The straightforward solution to this problem is performing reconstruction multiple times by traversing all possible candidates. However, this solution is extremely inefficient, which is not friendly for applications with computation resource constraints, especially when the number of candidates is large. To address this issue, we present our framework and provide the theoretical insights from convex optimization.

\section{Method}
\subsection{Preliminaries}
 We first adopt some general assumptions for IIPs similar to these works. Suppose a generative deep neural network is denoted $G$, which is often a non-convex function. 
Formally, the domain of the recovered signals is given by
    \begin{align}    
    \mathcal{S} = \{\boldsymbol{x} \in \mathbb{R}^n | \boldsymbol{x} = G(\boldsymbol{z};\boldsymbol{w})\},
    \end{align}
where $\boldsymbol{z}$ denotes the input of the model, which is often a fixed random number, and $\boldsymbol{w}$ denotes the weight of the neural network.
%When using generative prior, $\boldsymbol{w}$ is fixed, while $\boldsymbol{z}$ is fixed when using deep image prior.

\begin{assumption}
The ground truth signal $ \boldsymbol{x}_0$ belongs to the range of $G$ (i.e. the set of all potential outputs of $G$),
\begin{align}
    \boldsymbol{x}_0\in \mathcal{S}.
\end{align}
\end{assumption}
This assumption ensures the feasibility of recovering the original signal.

\begin{assumption}
$F$ is $\alpha$-strong convexity, $\beta$-strong smoothness w.r.t $\boldsymbol{x}$. This means for all $\boldsymbol{x}, \boldsymbol{x}' \in S$, $F$ satisfies,

% \begin{align}
% \frac{\alpha}{2} \| \boldsymbol{x} -\boldsymbol{y} \|_2^2 \leq F(\boldsymbol{y};\theta_i) - F(\boldsymbol{x};\theta_i) - \langle \nabla F(\boldsymbol{x};\theta_i), \boldsymbol{y} - \boldsymbol{x} \rangle, \\ \nonumber
% \frac{\beta}{2} \| \boldsymbol{x} - \boldsymbol{y} \|_2^2 \geq F(\boldsymbol{y};\theta_i) - F(\boldsymbol{x};\theta_i) - \langle \nabla F(\boldsymbol{x};\theta_i), \boldsymbol{y} - \boldsymbol{x} \rangle .
% \end{align}

\begin{align}
\frac{\alpha}{2} \| \boldsymbol{x} -\boldsymbol{x}' \|_2^2 \leq F(\boldsymbol{x}';\theta_i) - F(\boldsymbol{x};\theta_i) - \langle \nabla F(\boldsymbol{x};\theta_i), \boldsymbol{x}' - \boldsymbol{x} \rangle, \\ \nonumber
\frac{\beta}{2} \| \boldsymbol{x} - \boldsymbol{x}' \|_2^2 \geq F(\boldsymbol{x}';\theta_i) - F(\boldsymbol{x};\theta_i) - \langle \nabla F(\boldsymbol{x};\theta_i), \boldsymbol{x}' - \boldsymbol{x} \rangle .
\end{align}

\end{assumption}

The aforementioned works often use assumptions 1 and 2 to derive their theoretical guarantee under a known forward model's parameter. Hence, we make an assumption as,
\begin{assumption}
A signal $\boldsymbol{x}_0$ can be accurately reconstructed from its measurement with a known $\theta^*$ under a convergence guarantee if Assumptions 1 and 2 are fulfilled.
\end{assumption}
In our scenario, there is a set of candidate forward model parameters; therefore, we make an additional assumption to ensure the candidate set is reliable at least.
\begin{assumption}
There exists and only exists a $\epsilon$-suboptimal parameter $\hat{\theta}^*\in \Theta$, such that, 
\begin{align}
   \| \boldsymbol{y}-\mathcal{A}\left(\boldsymbol{x}_0;\hat{\theta}^*\right) \|_2^2 <\epsilon
\end{align}
 for a very small number $0<\epsilon<<1$.
\end{assumption}

\subsection{Moment-Aggregation Training Framework}

To solve IIPs under a set of candidate parameters, the idea is to construct a new loss $\mathcal{L}$  by using such a neural network $G$ presented in assumption 1. If the loss $\mathcal{L}$ satisfies the similar properties with $ F(\boldsymbol{x};\theta^*)$, the loss has a high probability of converging to a similar optimal with  $F(\boldsymbol{x};\theta^*)$. It is noteworthy that the neural network $G$ can only optimize $\boldsymbol{x}$ through optimizing $\boldsymbol{w}$ since $\theta_i$ can be viewed as an independent variable with $\boldsymbol{w}$. Now, we define the new loss and name it \textit{aggregation loss},
\begin{definition}
    Given a set of candidate parameters $\Theta$ and neural network $G$, any aggregation loss should satisfies: i) $\mathcal{L}$ is $\alpha$-strong convexity, $\beta$-strong smoothness w.r.t $\boldsymbol{x}$, and ii) $\lim_{\boldsymbol{x}\rightarrow \boldsymbol{x}_0} \mathcal{L}(\boldsymbol{x},\Theta)\rightarrow 0$.
\end{definition}
Here, the first condition ensures its convergence rate is tractable, while the second condition ensures the neural network can converge to the same optima as recovery by using the known precise parameter. 

Nevertheless, constructing a loss $\mathcal{L}$ is still challenging at this time because there is no prior knowledge about the quality of each candidate forward model parameter. Our solution is calculating the temporary quality of each candidate based on $F(\boldsymbol{x}; \theta_i)$ after every forward propagation of $G$. We define this time point as,
\begin{definition}
The \textbf{moment} is the time point between the forward propagation and backward propagation of each iteration by the neural network $G$.
\end{definition}
 Note that the surrogate qualities may not be super reliable at the beginning. i.e., $F(\boldsymbol{x}; \theta^*)\geq F(\boldsymbol{x}; \theta_i)$ is possible when the neural network does not converge well. 
However, with this surrogate quality of candidates, we are able to construct the \textit{moment-aggregation loss} (MA loss) that satisfies the conditions of \textit{aggregation loss} presented in definition 1 at each moment. We conjecture the loss in the entire lifetime should also have similar properties with \textit{aggregation loss} if each moment an MA loss has similar properties with  \textit{aggregation loss}

\begin{theorem}
    At each moment, a loss has the following format is an MA loss: 
    \begin{align}\label{eq:lossma}
        &\mathcal{L}_{MA}(\boldsymbol{x};\Theta) = \sum_{i=1}^{n_c}\omega_i F(\boldsymbol{x}; \theta_i),
    \end{align}
    where 
    \begin{align}\label{eq:weight}
        &\omega_i = H(i; F(\boldsymbol{x}; \theta_1),\cdots,F(\boldsymbol{x}; \theta_{n_c})  )\geq 0, \\ \nonumber
        & \textbf{Stop Gradient for }\omega_i,
    \end{align}
    Here, $H(i; F(\boldsymbol{x}; \theta_1),\cdots, F(\boldsymbol{x}; \theta_{n_c})  )$ is the function to calculate the weight for each candidate based on the surrogate qualities at each moment (i.e. $\omega_i$ will be updated at each iteration). Stopping gradient means when the neural network performs backward propagation, we consider every $\omega_i$ as a constant. $H$ should satisfy: i) $ \sum_{i=1}^{n_c}\omega_i =1$, and ii) $\lim_{\boldsymbol{x}\rightarrow \boldsymbol{x}_0}H(\theta^*; F(\boldsymbol{x}; \theta_1),\cdots,F(\boldsymbol{x}; \theta_{n_c}))\rightarrow 1$.

\end{theorem}
\begin{figure}[]
% \vspace{-0.4cm}
\centering
\includegraphics[width=0.99\columnwidth]{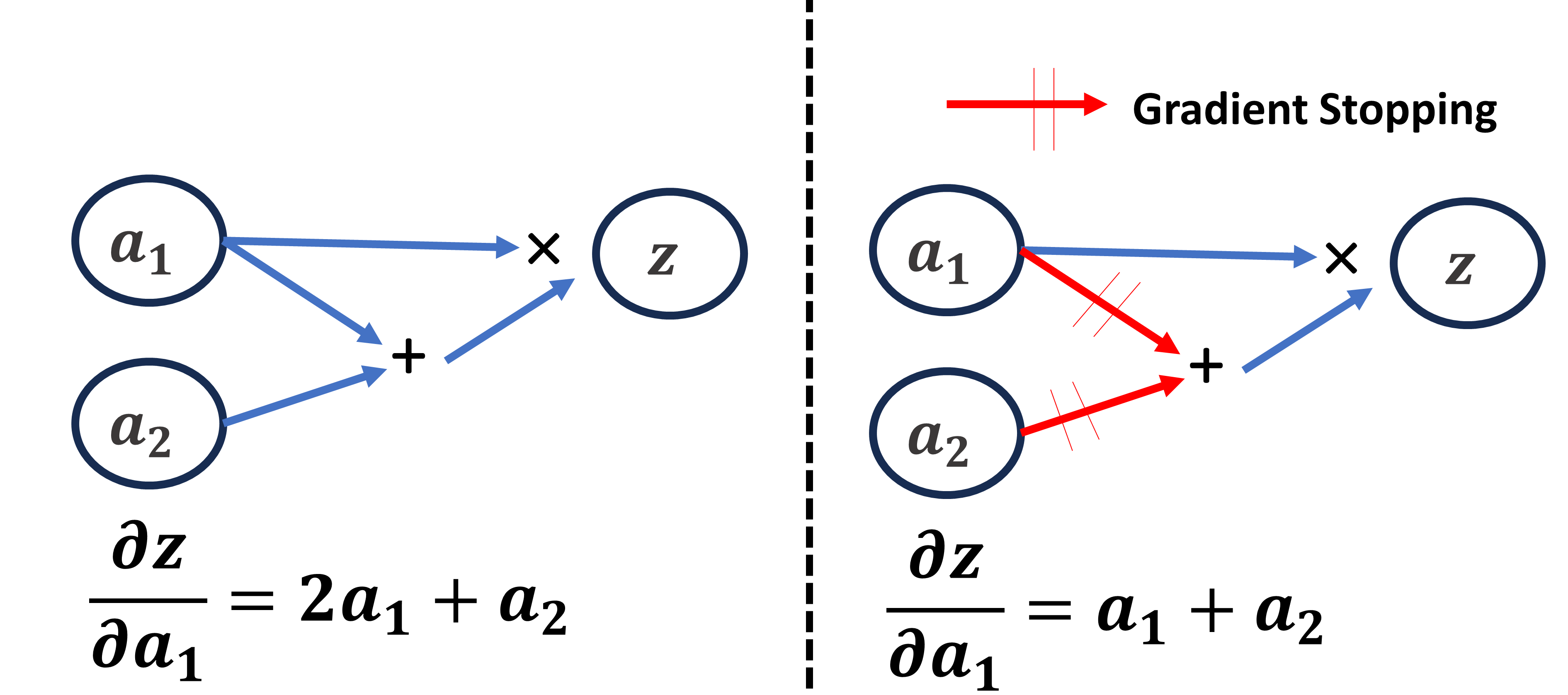} % Reduce the figure size so it is slightly narrower than the column. Don't use precise values for figure width. This setup will avoid overfull boxes.
%\vspace{-0.8cm}
\caption{\textbf{Left:} Derivative of $y$ w.r.t $a_1$ without gradient stopping. \textbf{Right:} Derivative with gradient stopping. }
\label{fig:gradientstopping}
% \vspace{-1cm}
\end{figure}
\begin{remark}
    It is noteworthy that the stop gradient plays a crucial role in the MA loss, since it preserves the convexity$/$smoothness by allowing us to use distributive law, i.e. $\nabla \mathcal{L}_{MA}(\boldsymbol{x};\Theta)=\nabla \sum_{i=1}^{n_c}\omega_i F_{\theta_i}(\boldsymbol{x}) $.
\end{remark}
 An example of how gradient stopping performs is shown in Fig. \ref{fig:gradientstopping}.

\begin{proof}
    First, we prove the convexity of $\mathcal{L}_{MA}$. For convenience, we denote $F(\boldsymbol{x}; \theta_i)$ as $F_{\theta_i}(\boldsymbol{x})$.  According to assumption 1,
    we easily obtain,
    %  \begin{align}\label{eq:11}
    %     &\omega_i\frac{\alpha}{2} \| \boldsymbol{x} -\boldsymbol{y} \|_2^2 \\ \nonumber
    %     &\leq \omega_i (F_{\theta_i}(\boldsymbol{y}) - F_{\theta_i}(\boldsymbol{x}) - \langle \nabla F_{\theta_i}(\boldsymbol{x}), \boldsymbol{y} - \boldsymbol{x} \rangle), \\ \nonumber
    %     &=\omega_i F_{\theta_i}(\boldsymbol{y}) - \omega_i F_{\theta_i}(\boldsymbol{x}) - \langle \omega_i\nabla F_{\theta_i}(\boldsymbol{x}), \boldsymbol{y} - \boldsymbol{x} \rangle, 
    % \end{align}
\begin{align}\label{eq:11}
        &\omega_i\frac{\alpha}{2} \| \boldsymbol{x} -\boldsymbol{x}' \|_2^2 \\ \nonumber
        &\leq \omega_i (F_{\theta_i}(\boldsymbol{x}') - F_{\theta_i}(\boldsymbol{x}) - \langle \nabla F_{\theta_i}(\boldsymbol{x}), \boldsymbol{x}' - \boldsymbol{x} \rangle), \\ \nonumber
        &=\omega_i F_{\theta_i}(\boldsymbol{x}') - \omega_i F_{\theta_i}(\boldsymbol{x}) - \langle \omega_i\nabla F_{\theta_i}(\boldsymbol{x}), \boldsymbol{x}' - \boldsymbol{x} \rangle, 
    \end{align}

Now we evaluate the convexity of $\mathcal{L}_{MA}$ presented in Eq. \ref{eq:lossma},
    \begin{align}\label{eq:121}
       &\mathcal{L}_{MA}(\boldsymbol{x}';\Theta)-\mathcal{L}_{MA}(\boldsymbol{x};\Theta)- \langle \nabla \mathcal{L}_{MA}(\boldsymbol{x};\Theta), \boldsymbol{x}' - \boldsymbol{x} \rangle  \\ \nonumber
       =& \sum_{i=1}^{n_c}\omega_i F_{\theta_i}(\boldsymbol{x}')-\sum_{i=1}^{n_c}\omega_i F_{\theta_i}(\boldsymbol{x})- \langle \nabla \sum_{i=1}^{n_c}\omega_i F_{\theta_i}(\boldsymbol{x}), \boldsymbol{x}' - \boldsymbol{x} \rangle \\ \nonumber
    =& \sum_{i=1}^{n_c}\omega_i(F_{\theta_i}(\boldsymbol{x}') - F_{\theta_i}(\boldsymbol{x}) - \langle \nabla F_{\theta_i}(\boldsymbol{x}), \boldsymbol{x}' - \boldsymbol{x} \rangle) \\ \nonumber
    \overset{(a)}{\geq}& \sum_{i=1}^{n_c}\omega_i\frac{\alpha}{2} \| \boldsymbol{x} -\boldsymbol{x}' \|_2^2 
    = \frac{\alpha}{2} \| \boldsymbol{x} -\boldsymbol{x}' \|_2^2. 
    \end{align}
Here, (a) in Eq. \ref{eq:121} is applying the inequality presented in Eq. \ref{eq:11}. Until now, the $\alpha$-convexity of $\mathcal{L}_{MA}$ is proved. Likewise, the $\beta$-strong smoothness can be proved. 

 Then, we prove $\mathcal{L}_{MA}$ that satisfies condition ii) in Definition 1. We substitute 
 \begin{align}
     \lim_{\boldsymbol{x}\rightarrow \boldsymbol{x}_0}H(\theta^*; F_{\theta_1}(\boldsymbol{x}),\cdots,F_{\theta_{n_c}}(\boldsymbol{x}))\rightarrow 1
 \end{align}
to Eq. \ref{eq:lossma},
\begin{align}
\lim_{\boldsymbol{x}\rightarrow \boldsymbol{x}_0}\mathcal{L}_{MA}(\boldsymbol{x};\Theta) \rightarrow F_{\theta^*}(\boldsymbol{x})
 \end{align}
Apparently, $\lim_{\boldsymbol{x}\rightarrow \boldsymbol{x}_0}F_{\theta^*}(\boldsymbol{x})\rightarrow 0$, hence the second condition is proved. The proof is completed.
\end{proof}

% \begin{theorem}
%   Under the assumptions, the MA loss will not reach any minima unless $\boldsymbol{x}\rightarrow \boldsymbol{x}_0$:
%   \begin{align}
%       \nabla \mathcal{L}_{MA}\neq 0, \forall \| \boldsymbol{x}-\boldsymbol{x}_0\|\geq \gamma
%   \end{align}
% \end{theorem}

% \end{corollary}

We propose one MA as 
$\omega_i= \frac{e^{1/F_{\theta_i}(\boldsymbol{x})}}{\sum_{j} e^{1/F_{\theta_j}(\boldsymbol{x})}}$.
    % \item \textbf{MA-2}: $\omega_i= \frac{1/F_{\theta_i}(\boldsymbol{x})}{\sum_{j} {1/F_{\theta_j}(\boldsymbol{x})}}$.
    % \item \textbf{MA-3}: $\omega_i= \mathbb{I}(F_{\theta_i}(\boldsymbol{x})= \min_j F_{\theta_j}(\boldsymbol{x}))$, where $j\in[n_c]$ and $\mathbb{I}$ is the indicator function.
 A summary of the training framework is presented in Algorithm 1.

\begin{algorithm}[ht]\label{alg:1}
% \scriptsize

\caption{ Moment-Aggregation Training Framework } 
\begin{algorithmic}[1]
\REQUIRE A neural network $G$, the set of candidate parameters $\Theta$, measurement $\boldsymbol{x}$, and a fixed number $\boldsymbol{z}$.
\ENSURE recovered signal $\boldsymbol{x}^*$.
\WHILE{not meet the stop criterion}
\STATE $\boldsymbol{x}\leftarrow G(\boldsymbol{z};\boldsymbol{w})$.
\STATE Compute $F_{\theta_i}$ for all $i\in [n_c]$ (Eq. \ref{eq:objective_me}).
\STATE Compute $\omega_i$ and \textbf{Stop Gradient} for $\omega_i$, for all $i\in [n_c]$ (Eq. \ref{eq:weight}).
\STATE Compute $\mathcal{L}_{MA}$ (Eq. \ref{eq:lossma}).
\STATE Update $\boldsymbol{w}$ via gradient-based optimization.
\ENDWHILE

\STATE $\boldsymbol{x}^*\leftarrow \boldsymbol{x}$.
\end{algorithmic}
\end{algorithm}

%(F(\boldsymbol{x}; \theta_i))\textbf{}
\section{Experiment}
We evaluate our proposed MA loss on two tasks: i) a standard CS problem and ii) a phase retrieval application.

\subsection{Standard CS problem}

\noindent\textbf{Setup.} We primarily evaluate our algorithm in the standard CS problem \cite{candes2006robust}, where $\mathcal{A}\left(\boldsymbol{x}_0;\theta^*\right)=\Phi\boldsymbol{x}_0$. The forward model parameter is a random Gaussian kernel $\Phi \in \mathbb{R}^{m \times n}$. Each element of $\Phi$ is Gaussian i.i.d and obeys $\Phi_{i, j} \sim \mathcal{N}\left(0, \frac{1}{m}\right)$. The set $\Theta$ consists 1 precise parameter and 9 candidate parameters randomly generated by the same distribution.
We choose two datasets for our evaluation: i) a toy dataset MNIST \cite{lecun1998gradient}, each image has $28\times 28$ pixels and ii) Shenzhen Chest X-Ray Dataset \cite{jaeger2014two}, we downsample each image to 256$\times$256 pixels. 
% We show most of the numerical results on these datasets. We also include:
% iii) retinopathy images from the STARE dataset \cite{hoover2000locating} with downsized to $128 \times 128$ pixels, and iv) drone-collected images from FLAME2 \cite{chen2022wildland} with downsized to $128 \times 128$ pixels.

% For grayscale we use the first 100 images in the test set of MNIST \cite{lecun1998gradient} and also 60 random images from the Shenzhen Chest X-Ray Dataset \cite{jaeger2014two}, downsampling a 512x512 crop to 256x256 pixels. For RGB we use retinopathy images from the STARE dataset \cite{hoover2000locating} with $512 \times 512$ crops downsized to $128 \times 128$ pixels.

\begin{table*}[]
\centering
\caption{Comparison of reconstruction performance by different methods.}
 \label{tab:1}
\resizebox{0.65\textwidth}{!}{%
\begin{tabular}{cccccc}\toprule
 & \textbf{Dataset} & \multicolumn{2}{c}{\textbf{MNIST}} & \multicolumn{2}{c}{\textbf{X-ray}} \\ 
\multirow{-2}{*}{\textbf{Method}} & \textbf{$m$} & \textbf{100} & \textbf{200} & \textbf{1000} & \textbf{2000} \\ \midrule
 & PSNR & {\color[HTML]{C0C0C0} 10.224} & {\color[HTML]{C0C0C0} 10.604} & {\color[HTML]{C0C0C0} 7.733} & {\color[HTML]{C0C0C0} 8.274} \\
\multirow{-2}{*}{\begin{tabular}[c]{@{}c@{}}Random Parameter\\      (lasso-wavelet)\end{tabular}} & SSIM & {\color[HTML]{C0C0C0} 0.215} & {\color[HTML]{C0C0C0} 0.255} & {\color[HTML]{C0C0C0} 0.006} & {\color[HTML]{C0C0C0} 0.005} \\ \midrule
 & PSNR & {\color[HTML]{C0C0C0} 10.224} & {\color[HTML]{C0C0C0} 10.607} & {\color[HTML]{C0C0C0} 7.376} & {\color[HTML]{C0C0C0} 8.036} \\
\multirow{-2}{*}{\begin{tabular}[c]{@{}c@{}}Random Parameter\\      (lasso-DCT)\end{tabular}} & SSIM & {\color[HTML]{C0C0C0} 0.215} & {\color[HTML]{C0C0C0} 0.255} & {\color[HTML]{C0C0C0} 0.004} & {\color[HTML]{C0C0C0} 0.001} \\ \midrule
 & PSNR & {\color[HTML]{C0C0C0} 10.223} & {\color[HTML]{C0C0C0} 10.606} & {\color[HTML]{C0C0C0} 7.501} & {\color[HTML]{C0C0C0} 8.002} \\
\multirow{-2}{*}{\begin{tabular}[c]{@{}c@{}}Random Parameter\\      (BM3D-AMP)\end{tabular}} & SSIM & {\color[HTML]{C0C0C0} 0.214} & {\color[HTML]{C0C0C0} 0.250} & {\color[HTML]{C0C0C0} 0.004} & {\color[HTML]{C0C0C0} 0.003} \\ \midrule
 & PSNR & {\color[HTML]{C0C0C0} 10.212} & {\color[HTML]{C0C0C0} 10.601} & {\color[HTML]{C0C0C0} 7.533} & {\color[HTML]{C0C0C0} 8.675} \\
\multirow{-2}{*}{\begin{tabular}[c]{@{}c@{}}Random Parameter\\      (CS-DIP)\end{tabular}} & SSIM & {\color[HTML]{C0C0C0} 0.204} & {\color[HTML]{C0C0C0} 0.247} & {\color[HTML]{C0C0C0} 0.005} & {\color[HTML]{C0C0C0} 0.007} \\ \midrule
 & PSNR & {\color[HTML]{C0C0C0} 10.084} & {\color[HTML]{C0C0C0} 11.004} & {\color[HTML]{C0C0C0} 7.813} & {\color[HTML]{C0C0C0} 9.215} \\
\multirow{-2}{*}{Uniform Aggregation} & SSIM & {\color[HTML]{C0C0C0} 0.224} & {\color[HTML]{C0C0C0} 0.295} & {\color[HTML]{C0C0C0} 0.006} & {\color[HTML]{C0C0C0} 0.009} \\ \midrule
 & PSNR & 10.221 & 13.301 & 19.163 & 19.949 \\
\multirow{-2}{*}{Alternating} & SSIM & 0.262 & 0.457 & 0.330 & 0.381 \\ \midrule
 & PSNR & 15.542 & 19.464 & 23.669 & 25.081 \\
\multirow{-2}{*}{Upper bound} & SSIM & 0.620 & 0.801 & 0.568 & 0.638 \\ \midrule
 & PSNR & 15.204 & 18.293 & 22.051 & 23.141 \\
 & $\Delta$ & 0.337 & 1.171 & 1.618 & 1.940 \\
 & SSIM & 0.580 & 0.732 & 0.505 & 0.567 \\
\multirow{-4}{*}{\textbf{Ours}} & $\Delta$ & 0.041 & 0.069 & 0.063 & 0.071 \\ \bottomrule
\end{tabular}%
}
\end{table*}

\noindent\textbf{Implementation.} Our experiment is based on the unlearned training pipeline \textit{CS-DIP} provided by \cite{van2018compressed}. The neural network is the generator of DCGAN \cite{radford2015unsupervised} and is randomly initialized. We use Adam optimizer \cite{kingma2014adam} with a fixed learning rate 1e-3. The experiments are conducted on a cluster node with a V100 16G GPU.

\noindent\textbf{Baselines.} We include: i) \textbf{Upper-bound}: The most important baseline is the reconstruction with the known precise parameter, which is treated as the upper bound of our problem. ii) \textbf{Random Parameter}: We random select a parameter from the set $\Theta$. This is the blind reconstruction, and we simply compute the expected value of reconstruction results by using every candidate. We include Lasso-wavelet, lasso-DCT, BM3D-AMP \cite{metzler2016denoising}, and CS-DIP \cite{van2018compressed}.
iii) \textbf{Uniform Aggregation}: We consider every candidate parameter to have the same quality (i.e. $w_i=1/n_c$). and iv) \textbf{Alternating Optimization}: In each epoch, this baseline involves first updating the neural network, then finding a good $\theta_i$ that has the minimum loss and backpropagating this loss. 

\noindent\textbf{Evaluation Metrics.} We employ two widely used metrics to measure the reconstruction performance with ground truth: i) Peak Signal-to-Noise Ratio (PSNR) and ii) Structural Similarity Index Measure (SSIM).

% \begin{figure}[!t]
% % \vspace{-0.4cm}
% \centering
% \includegraphics[width=0.49\textwidth]{vis_mnist.jpg} % Reduce the figure size so it is slightly narrower than the column. Don't use precise values for figure width. This setup will avoid overfull boxes.
% \caption{Comparison of different reconstruction strategies in MNIST dataset when the $m$ is 200.}
% \label{fig:vismnist}
% % \vspace{-1cm}
% \end{figure}

% \begin{figure}[!t]
% % \vspace{-0.4cm}
% \centering
% \includegraphics[width=0.49\textwidth]{vis_Xray.jpg} % Reduce the figure size so it is slightly narrower than the column. Don't use precise values for figure width. This setup will avoid overfull boxes.
% \caption{Comparison of different reconstruction strategies in X-ray dataset when $m$ is 2000 (total pixel is 65536)}
% \label{fig:visxray}
% % \vspace{-1cm}
% \end{figure}

\begin{figure*}[!t]
% \vspace{-0.4cm}
\centering
\includegraphics[width=0.99\textwidth]{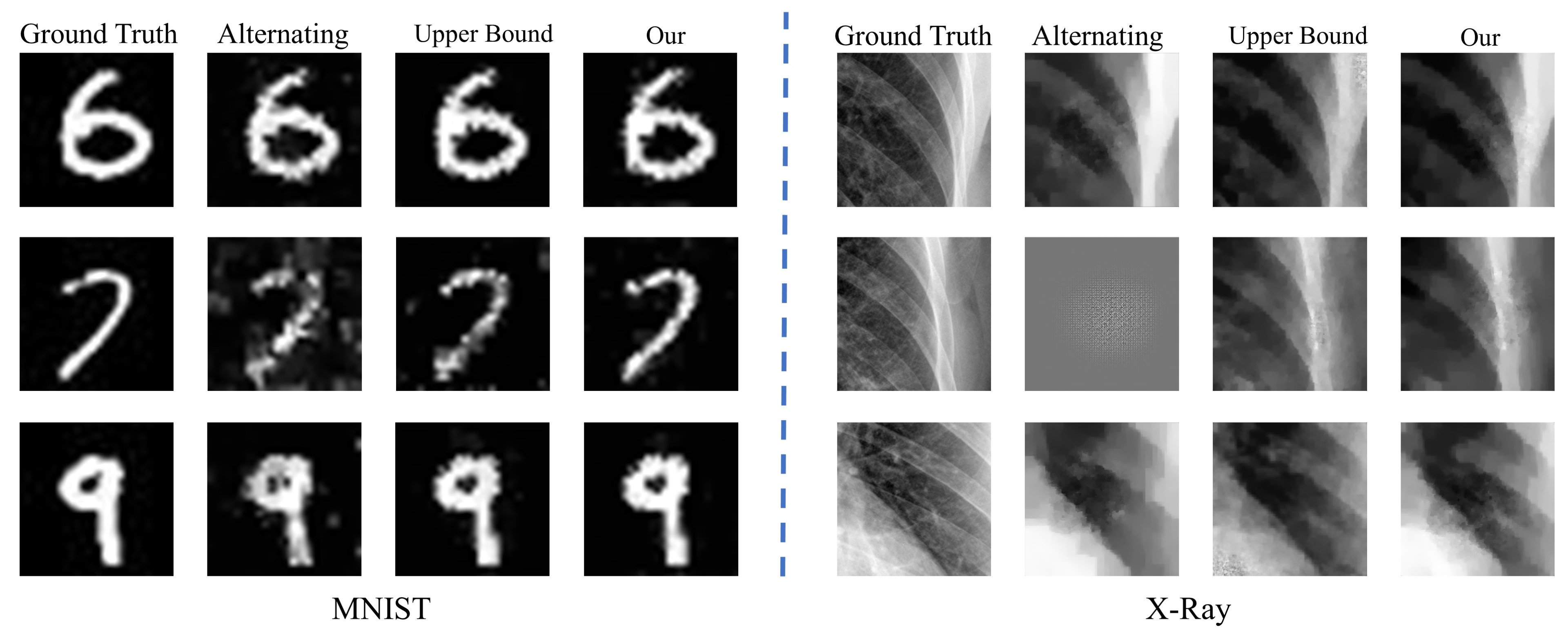} % Reduce the figure size so it is slightly narrower than the column. Don't use precise values for figure width. This setup will avoid overfull boxes.
\caption{ Comparison of different reconstruction strategies in \textbf{Left:} MNIST dataset when the $m$ is 200. \textbf{Right:} X-ray dataset when $m$ is 2000 (total pixel is 65536).}
\label{fig:vismnist}
% \vspace{-1cm}
\end{figure*}

\begin{figure*}[]
%\vspace{-0.4cm}
\centering
\includegraphics[width=0.99\textwidth]{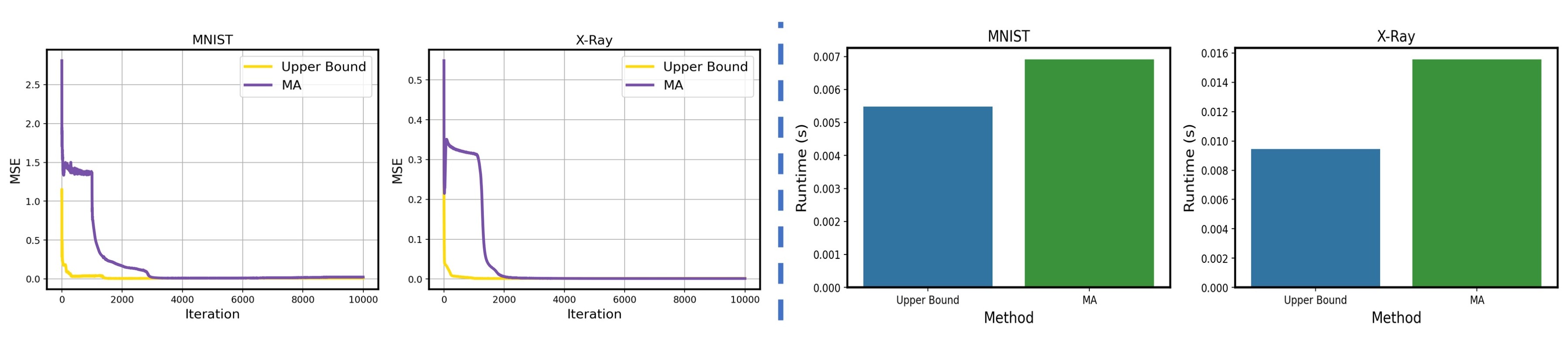} % Reduce the figure size so it is slightly narrower than the column. Don't use precise values for figure width. This setup will avoid overfull boxes.
%\vspace{-0.1cm}
\caption{Comparison of the upper bound and our proposed method (MA) for \textbf{Left:} convergence and \textbf{Right:} runtime each iteration.i.e. forward-propagation and backward-propagation. }
\label{fig:convergence}
% \vspace{-1cm}
\end{figure*}

\noindent\textbf{Results.} The numerical results are shown in Table \ref{tab:1}. The first observation is that using Random Parameter to reconstruct the signal blindly is not feasible, which only achieves around 10dB PSNR, meaning almost nothing is reconstructed. This is consistent with the fundamental principle of IIPs. Then, we observe our methods can achieve similar reconstruction results with the upper bound, which is reconstructing using the known parameter. For example, in both MNIST and X-ray datasets, our method only has a 0.04-0.07 SSIM reduction.   
Another interesting observation is that alternating optimization shows obvious superiority over blind reconstruction, and it can reconstruct the signal sometimes, e.g. around 19 dB in PSNR for X-ray image reconstruction. However, this method is not stable since it quickly switches different candidate parameters to optimize, which results in a high probability of failure to reconstruct. Some samples of reconstructed signals for MNIST and X-ray datasets are shown in Fig. \ref{fig:vismnist}(\textbf{Left} and \textbf{Right}), respectively. Both of them illustrate that reconstructed signals by our method can achieve very similar performance with the upper bound. We also demonstrate the convergence rate in Fig.\ref{fig:convergence}(\textbf{Left}), which shows our method can converge to the same level of reconstruction error with a 
lagging. This lagging is reasonable, because the upper bound using the known precise parameter is easy to converge, while under an uncertain set of candidates, the error landscape for optimization is more complicated. Fig.\ref{fig:convergence}(\textbf{Right}) shows the runtime for each epoch by using a set of candidate parameters and only one precise parameter. Our method's overhead is caused by the computation of the forward process for each candidate parameter. Although the overhead exists, our method is still much faster than training different neural networks separately with different candidate parameters. For example, if we only have one device that can train the model, in the MNIST dataset, our method requires $0.007$ seconds to update for every epoch; however, training 10 different neural networks for different candidates requires approximately $0.005\times 10 = 0.5$ seconds.

\subsection{Applications in Phase Retrieval}
\noindent\textbf{Setup.} We also show the feasibility of our methods in phase retrieval. Here, we take holographic imaging as an example \cite{zhang2018twin}. Suppose $\boldsymbol{O}(d=0)$ denotes a complex-valued object wave at location $d=0$. We can use the angular spectrum method to describe the propagation of the wave to the sensor plane $d=z$ as $\boldsymbol{O}(d=z)= \mathcal{F}^{-1}\{\boldsymbol{P}(\lambda,d=z)\cdot\mathcal{F}\{\boldsymbol{O}(x,y;d=0)\}\}$, where $\lambda$ denotes the wavelength, $x,y$ denotes the coordinate in the object space that is orthogonal with $z$, and $\mathcal{F}$ and $\mathcal{F}^{-1}$ denote Fourier transform and inverse Fourier transform, respectively. $\boldsymbol{P}(\lambda,d=z)$ is called the transfer function and is based on the experiment equipment and setups (Refer to \cite{zhang2018twin}). Similarly, a reference plane wave can propagate to the sensor plane. The sensor plane captures the superposition of the object wave and reference wave as $\boldsymbol{H}=\|\boldsymbol{O}^2(d=z)+\boldsymbol{R}^2(d=z) \|^2$, known as a hologram, and our goal is to retrieve $\boldsymbol{O}(d=0)$ from $\boldsymbol{H}$. This problem is also an ill-posed IIP problem, and here $\boldsymbol{P}(\lambda,d=z)$ can be considered as the forward model with uncertain parameters due to the low-quality equipment or an inaccurate precision optical rail.
In our simulation, we set the known wavelength and distance to $\lambda=0.520\mu m$ and  $5000\mu m$ to generate holograms, respectively. The set of uncertain parameters $\Theta=\{d_1,\cdots,d_{10}\}$ is generated by $d_i\sim \mathcal{U}(z-500,z+500)$. We choose samples from the Gland segmentation dataset (GlaS)~\cite{GlaS} and the Multi-Organ Nucleus Segmentation (MoNuSeg) dataset~\cite{MoNuSeg} to generate the simulated holograms.

\noindent\textbf{Baseline.} i) \textbf{Upper-bound}: the reconstruction with a known forward model parameter. ii) \textbf{Random Parameter}: We evaluate CS-DIP in this application, which presented in \cite{niknam2021holographic}.
iii) \textbf{Uniform Aggregation}, and iv) \textbf{Alternating Optimization}. 

\noindent\textbf{Evaluation Metrics.} PSNR, and SSIM.
% We perform the same evaluation metric as the Standard CS problem, including PSNR, and SSIM.
% \usepackage{multirow}
% \usepackage{graphicx}

\begin{figure*}[!t]
%\vspace{-0.4cm}
\centering
\includegraphics[width=0.99\textwidth]{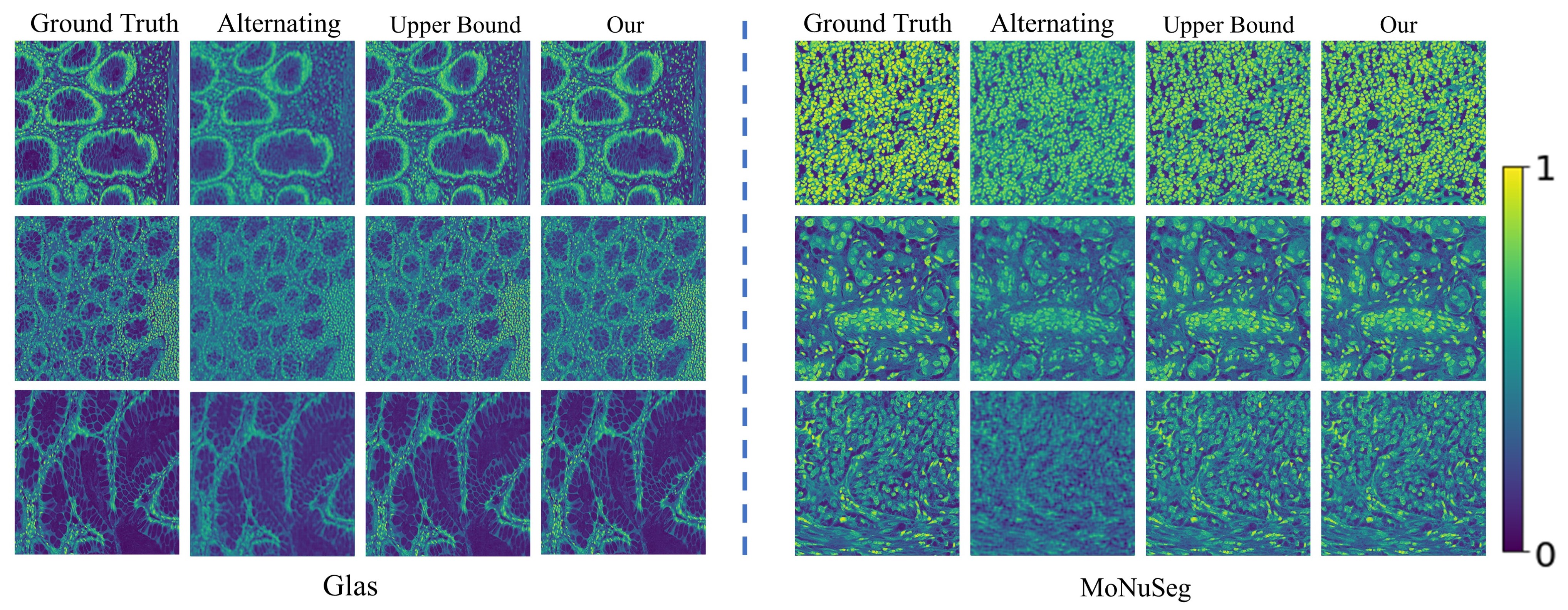} % Reduce the figure size so it is slightly narrower than the column. Don't use precise values for figure width. This setup will avoid overfull boxes.
\caption{The reconstructed phase by different methods based on Glas and MoNuSeg dataset. Please \textcolor{red}{\textbf{zoom in}} for a detailed comparison.}
\label{fig:vis_PHASE}
% \vspace{-1cm}
\end{figure*}

\begin{table}[]
\centering
\caption{Comparison of reconstruction performance by different methods in the application of phase retrieval.}
\label{tab:2}
\resizebox{0.45\textwidth}{!}{%
\begin{tabular}{cccc}\toprule
\multicolumn{1}{l}{\textbf{Method}} & \textbf{Dataset} & \textbf{Glas} & \textbf{MoNuSeg} \\ \midrule
\multirow{2}{*}{Random Parameter} & PSNR & {\color[HTML]{C0C0C0} 19.8782 } & {\color[HTML]{C0C0C0} 18.190} \\ %PSNR: 18.19011706282526 |  SSIM: 0.6715107073621239
 & SSIM & {\color[HTML]{C0C0C0} 0.6715} & {\color[HTML]{C0C0C0}  0.584} \\ \midrule
\multirow{2}{*}{Uniform Aggregation} & PSNR & {\color[HTML]{C0C0C0} 19.626} & {\color[HTML]{C0C0C0} 19.107} \\
 & SSIM & {\color[HTML]{C0C0C0} 0.579} & {\color[HTML]{C0C0C0} 0.549} \\ \midrule
\multirow{2}{*}{Alternating} & PSNR & 19.903 & 18.110 \\
 & SSIM & 0.685 & 0.530 \\ \midrule
\multirow{2}{*}{Upper bound} & PSNR & 28.519 & 25.392 \\
 & SSIM & 0.959 & 0.934 \\ \midrule
\multirow{4}{*}{\textbf{ours}} & PSNR & 28.212 & 25.225 \\
 & $\Delta$ & 0.307 & 0.167 \\
 & SSIM & 0.941 & 0.931 \\
 & $\Delta$ & 0.018 & 0.003 \\ \bottomrule
\end{tabular}%
}
\end{table}

\noindent\textbf{Results.} The results are shown in Table \ref{tab:2} and Fig. \ref{fig:vis_PHASE}. In this experiment, our method consistently demonstrates a small gap in the reconstruction with the known precise parameter. For example, there are 0.307 and 0.167 gaps in PNSR for Glas and MoNuSeg, respectively. We also find that using Random 
 Parameter, Uniform Aggregation and Alternating Optimization can reconstruct the low-frequency information of the object (e.g. outline and shape, as shown in the second column of Fig. \ref{fig:vis_PHASE} for Alternating Optimization) while lacking the reconstruction of the detailed texture. This may be because the reproduced measurement (i.e. $\hat{\boldsymbol{y}}$ the prediction after the forward process) in this task is still like an image, which can be partially fitted by the neural network. However, the detailed texture represents depth information, which is crucial in this task; hence, these methods are considered to fail to reconstruct the signal in this sense.
 
 % alternating optimization involves quickly switching different candidate measurements to optimize, which leads to a large gradient change and results in unstable training.

\section{Discussion and Conclusion}

This paper focuses on a scenario addressing inverse imaging problems (IIPs), where the main challenge arises from uncertainties in the parameters of the forward model used for the imaging process. These uncertainties can stem from various sources, such as calibration drifts in imaging devices, imprecise knowledge of the device parameters, or variations in experimental setups, making the task of reconstructing the original image from its compressed measurements particularly difficult. In this work, we consider there are a set of candidate parameters. Instead of testing different candidate parameters independently, our proposed MA framework marks a significant step forward under this parameter uncertainty by effectively aggregating information from all candidate parameters of the forward model. Our theoretical analysis is built on the aforementioned works, where they provide the convergence guarantee under the assumption that the forward model parameter is known. We take a step forward to show that we can construct a loss under a set of candidate parameters with similar properties to the loss with a known parameter, and hence, convergence by our method is ensured.  Our experimental results demonstrate that the MA framework achieves a close performance to that of reconstructions using known forward model parameters(upper bound). Specifically, our method only has a 0.04-0.07 SSIM difference with the upper bound in MNIST and X-ray dataset, respectively. Additionally, there are only 0.307 and 0.167 reductions in PNSR for the Glas and MoNuSeg datasets, respectively. 

This proposed method demonstrates significant potential in scenarios where accurate parameters remain unknown, particularly in medical imaging, including fundus camera imaging, microscopic imaging, MRI, and CT.
We admit performance gaps and occasionally unstable reconstruction still exist, and we conjecture this is because of the complicated error landscape in real optimization beyond our assumptions, which will be investigated in the future. Future work will explore extending the MA framework to more complex imaging models and closer to real-world scenarios.
\vspace{-0.1cm}
\section*{Acknowledgments}
This material is based upon the work supported by the National Science Foundation under Grant Number CNS-2204721 and the MIT Lincoln Laboratory under Grant numbers 2015887 and 7000612889.
% \noindent\textbf{ACKNOWLEDGMENTS}
% The work was partially supported by .

% our method demonstrates not only theoretical robustness but also practical efficacy across different datasets and problem setups. 

% In this paper, we focus on scenarios where the parameters of the forward model are uncertain or partially unknown, which is a general issue in broad applications that require a parameterized forward process. The reconstruction results can be significant 
% the measurement device may undergo calibration drifts

% In this paper, we consider the inverse problem under a forward model with a set of candidate parameters, which is a general issue in broad applications that require a parameterized forward process.

% In real-world application scenarios, where the optimal parameters cannot be determined, but a set of candidate parameters exists, existing methods typically consider each set of parameters independently.

% Instead of testing different candidate parameters independently, our proposed MA framework can consider the effect from all candidates simultaneously to reconstruct the signal. The experimental results illustrate our method can achieve a similar reconstruction with it using the known parameter. 

\newpage

{
    \small
    \bibliographystyle{ieeenat_fullname}
    \bibliography{main}
}


\end{document}